\documentclass{article}

\usepackage{PRIMEarxiv}
\usepackage{xcolor}
\usepackage{array}
\usepackage{booktabs}
\usepackage{colortbl}
\usepackage{multicol}
\usepackage{graphicx}
\usepackage{arydshln}
\usepackage{multirow}
\usepackage[utf8]{inputenc} 
\usepackage[T1]{fontenc}    
\usepackage{hyperref}       
\usepackage{url}            
\usepackage{booktabs}       
\usepackage{amsfonts}       
\usepackage{nicefrac}       
\usepackage{microtype}      
\usepackage{lipsum}
\usepackage{fancyhdr}       
\usepackage{graphicx}       
\graphicspath{{media/}}     

\usepackage[utf8]{inputenc}
\usepackage{rotating}
\usepackage{multirow}
\usepackage{longtable}
\usepackage{colortbl}

\newcommand{\etal}{\textit{et al.}}

\title{Translating Radiology Reports into Plain Language using ChatGPT and GPT-4 with Prompt Learning: Promising Results, Limitations, and Potential
}

\author{
  Qing Lyu, Josh Tan, Michael E. Zapadka, Janardhana Ponnatapura, Christopher T. Whitlow \\
  Wake Forest University School of Medicine \\
  Winston-Salem, NC \\
  \texttt{\{qlyu, jtan, mzapadka, jponnata, cwhitlow\}@wakehealth.edu} \\
   \And
  Chuang Niu, Ge Wang \\
  Rensselaer Polytechnic Institute \\
  Troy, NY\\
  \texttt{\{niuc, wangg6\}@rpi.edu} \\
  \And
  Kyle J. Myers \\
  Puente Solutions LLC \\
  Phoenix, AZ \\
  \texttt{drkylejmyers@gmail.com}
}

\begin{document}
\maketitle

\begin{abstract}
The large language model called ChatGPT has drawn extensively attention because of its human-like expression and reasoning abilities. In this study, we investigate the feasibility of using ChatGPT in experiments on using ChatGPT to translate radiology reports into plain language for patients and healthcare providers so that they are educated for improved healthcare. Radiology reports from 62 low-dose chest CT lung cancer screening scans and 76 brain MRI metastases screening scans were collected in the first half of February for this study. According to the evaluation by radiologists, ChatGPT can successfully translate radiology reports into plain language with an average score of 4.27 in the five-point system with 0.08 places of information missing and 0.07 places of misinformation. In terms of the  suggestions provided by ChatGPT, they are general relevant such as keeping following-up with doctors and closely monitoring any symptoms, and for about 37\% of 138 cases in total ChatGPT offers specific suggestions based on findings in the report. ChatGPT also presents some randomness in its responses with occasionally over-simplified or neglected information, which can be mitigated using a more detailed prompt. Furthermore, ChatGPT results are compared with a newly released large model GPT-4, showing that GPT-4 can significantly improve the quality of translated reports. Our results show that it is feasible to utilize large language models in clinical education, and further efforts are needed to address limitations and maximize their potential. 
\end{abstract}


\section{Introduction}
Since OpenAI released ChatGPT, a state-of-the-art natural language processing (NLP) model in November 2022, ChatGPT has received global attention with over 100 million users because of its human-like expression and reasoning abilities \cite{reuters, theguardian}. ChatGPT answers users' general queries as if it is a human and performs various tasks from poem composition, essay writing, as well as coding including debugging. Compared with the previous NLP models like BERT \cite{devlin2018bert}, XLNet \cite{yang2019xlnet}, and GPT \cite{radford2018improving}, ChatGPT is a quantum leap characterized by several characteristic features: a larger model with more parameters, chain of thought prompting, and trained with reinforcement learning from human feedback (RLHF). ChatGPT was developed based on GPT-3 which has 175 billion parameters and the aforementioned other models which have with less than 200 million parameters. Prompting learning is used to induce the reasoning process effectively. RLHF injects high-quality human knowledge and helps align the results of ChatGPT to be friendly and safe to the society \cite{ouyang2022training}. 

Given the huge success of ChatGPT, recently there are studies on adapting ChatGPT for downstream tasks, like writing a systematic literature review~\cite{wang2023can}, medical school education~\cite{kung2023performance}, language translation~\cite{liebrenz2023generating}, scholar content generation for publication~\cite{liebrenz2023generating}, and solving mathematical problems~\cite{ patel2023chatgpt}. In addition to these interesting studies, the investigation of ChatGPT for clinical usage has been extensively studied. Patel~\etal attempted to use ChatGPT to write a patient discharge summary and discussed some concerns~\cite{patel2023chatgpt}. Biswas explored using ChatGPT for medical writing applications, from patient care-related writing, medical publication, medical administrative documentation, and meeting summarization~\cite{biswas2023chatgpt}. Jeblick~\etal investigated the quality of ChatGPT's radiology report simplification, and concluded that the simplified reports were factually correct, complete, and not harmful
to the patient~\cite{jeblick2022chatgpt}. Rao~\etal demonstrated the feasibility of using ChatGPT as an adjunct for radiology decision making~\cite{rao2023evaluating}. Sarraju~\etal used ChatGPT to provide cardiovascular disease prevention recommendations and found over 80\% of ChatGPT's responses were appropriate~\cite{sarraju2023appropriateness}.

Radiology reports summarize experts' opinions on medical images acquired with radiography, computed tomography (CT), magnetic resonance imaging (MRI), nuclear, ultrasound and optical imaging methods. Findings in these reports are instrumental for diagnosis and treatment. However, there are often too many medical terminologies in such a report that are difficult to be understood for patients without any medical  background. With ChatGPT, it is now possible to re-express a professional report in plain language so that patients know what their radiology reports mean, which will be invaluable to reduce anxiety, promote compliance, and improve outcomes. 

In this study, we focus on the performance of ChatGPT in translating radiology reports into layman versions. Also, we ask ChatGPT to provide suggestions for both patients and healthcare providers based on each radiology report, and then we evaluate the quality of provided suggestions. Furthermore, we compare the results of ChatGPT with that obtained using the newly released GPT-4.

\section{Methodology}

\subsection{Report acquisition}
To show the performance of ChatGPT on a set of representative radiology reports, we collected 62 chest CT screening reports and 76 brain MRI screening reports from the Atrium Health Wake Forest Baptist clinical database. All reports were generated between February $1^{st}$ and $13^{th}$. All reports were de-identified by removing sensitive patient information.

The chest CT screening reports followed the low-dose chest CT lung cancer screening protocol without involving contrast agents. The patient are between 53 and 80 years old with an average of 66.9 years old (32 male and 30 female). The reports were finalized by 11 experienced radiologists with an average of $278 \pm 57$ words. The 
reports were classified into 6 classes based on the overall Lung-RADS category shown in each report (1, 1S, 2, 2S, 3, 4A).

\begin{table}[!t]
\caption{Statistics of the chest CT screening reports. \label{table1}}%
\centering
\begin{tabular}{c | c c c c c c | c}
\toprule
 & 1 & 1S & 2 & 2S & 3 & 4A & overall\\
\midrule
Category Count & 15 & 2 & 35 & 5 & 1 & 4 & 62 \\
Category Percentage & 24\% & 3\% & 56\% & 8\% & 2\% & 6\% & 100\% \\
Age (year) & 65.2$\pm$6.2 & 62.5$\pm$6.4 & 67.3$\pm$6.0 & 71.4$\pm$5.6 & 60.0$\pm$0 & 69.0$\pm$8.8 & 66.9$\pm$6.3 \\
\bottomrule
\end{tabular}
\end{table}

The brain MRI screening reports followed the brain tumor protocol with and without the use of contrast agent. The patient age range is between 5 and 98 with an average of 55.0 years old (45 male and 31 female). The reports were finalized by 14 experienced radiologists with $247 \pm 92$ words. Reports were classified into 3 classes based on the findings on metastases: no metastases, stable condition without newly emerging or growing metastases, and worsening condition with growing or newly emerging metastases.

\begin{table}[!t]
\caption{Statistics of the brain MRI screening reports.\label{table2}}%
\centering
\begin{tabular}{c | c c c | c}
\toprule
 & no mats & stable & worsening & overall\\
\midrule
Category Count & 11 & 40 & 25 & 76 \\
Percentage & 14\% & 53\% & 33\% & 100\% \\
Age (year) & 63.6$\pm$11.5 & 47.9$\pm$21.0 & 60.9$\pm$16.8 & 54.5$\pm$19.7 \\
\bottomrule
\end{tabular}
\end{table}

\subsection{Experimental design}
In our experiments, we gave ChatGPT the following three prompts and recorded its responses:
\begin{itemize}
  \item Please translate a radiology report into plain language that is easy to understand.
  \item Please provide some suggestions for the patient.
  \item Please provide some suggestions for the healthcare provider.  
\end{itemize}
All the ChatGPT responses were collected in the mid February.

\subsection{Performance evaluation}
After collecting all ChatGPT responses, we invited two experienced radiologists (21 and 8 years of experiences) to evaluate the quality of the ChatGPT responses. 

For the report translation, the evaluation effort was focused on the three aspects: overall score, completeness, and correctness. The 
radiologists recorded the number of places where information is missed and also the number of places for incorrect information in each of the translated reports, and gave an overall score based on the 5-point system (1 for worst and 5 for best). We then conducted statistical analysis on radiologists' feedback. For example, if there are ten translated reports and radiologists found one place of information missing among them, we would conclude that there are an average 0.1 places of information missing.

In terms of the suggestion evaluation, statistical analysis was performed to record high-frequency suggestions, the percentage of specific suggestions on a certain finding in the report, and the percentage of inappropriate suggestions that are not related to any finding in the report. 

\section{Key results}
\subsection{ChatGPT-translated reports versus the original reports}
Compared with the original radiology reports, ChatGPT generated plain language 
versions with generally fewer words in both chest CT and brain MRI cases. For the chest CT reports, there are 85.5\% translation results (53 of 62) are shorter than the corresponding original reports with an overall length reduction of 26.7\%. Specifically, ChatGPT can reduce the length of the original reports by 20.5\%, 29.0\%, 29.0\%, 54\%, and 29.4\% for Lung-RADS category 1, 2, 2S, 3, and 4A respectively. The only exception is the 1S category with a length increment of 13.3\% after ChatGPT translation. For the brain MRI radiology reports, 72.4\% of the translation results (55 of 76) have fewer words than corresponding original reports with an overall length reduction of 21.1\%. Except for the "no mats" category with slightly (1.8\%) more words, the reports in all the other categories are shorter after ChatGPT translation. Specifically, the plain language versions of reports in "stable" and "worsening" categories are respectively 13.1\% and 34.1\% shorter than the original versions. 

A typical scenario of paragraph shortening happens when there were multiple places in the radiology report showed no abnormality. Then, ChatGPT summarized all those negative findings together in a single sentence. For example, in a chest CT report, it said "PLEURA: No pleural thickening or effusion. No pneumothorax. HEART: Heart size normal. No pericardial effusion. CORONARY ARTERY CALCIFICATION: None. MEDIASTINUM/HILUM/AXILLA: No adenopathy." ChatGPT translated the text into "The pleura, heart and blood vessels are normal, and there is no sign of cancer in the lymph nodes."

Apart from shortening paragraphs and distilling information, the translated reports are patient-friendly and easier to understand by replacing medical terminologies with common words. For example, in a chest CT report regarding the findings in lungs stated \textit{"Granuloma seen in the right middle lobe 1mm."} ChatGPT translated the  text into the following sentence: \textit{"There is a small 1mm area in the right middle lobe that looks like a granuloma, which is a small area of inflammation that is usually not concerning."} After ChatGPT translation, the medical terminology of granuloma was explained with its severity also mentioned. 

Another great character of the translated report is information integration. ChatGPT is capable of integrating information shown in different sections of the original  report  so that the patient can better understand the report. A good example is a chest CT report. This report was compared with the scan conducted on August 6, 2021 in the comparison section. In the findings section, there is a sentence \textit{"There is a right lower lobe granuloma 6 mm unchanged."}. ChatGPT integrated information shown in the comparison and findings sections, and generated the following sentence: \textit{"There is also a 6mm granuloma in the right lower lobe, but it has not changed since a previous CT scan done in August 2021."}

\begin{table}[!t]
\caption{Comparison of the Chest CT screening reports and their ChatGPT translations. \label{table3}}%
\centering
\begin{tabular}{>{\centering\arraybackslash}m{2.6cm}|>{\centering\arraybackslash}m{1.5cm}>{\centering\arraybackslash}m{1.5cm}>{\centering\arraybackslash}m{1.5cm}>{\centering\arraybackslash}m{1.5cm}>{\centering\arraybackslash}m{1.2cm}>{\centering\arraybackslash}m{1.5cm}|>{\centering\arraybackslash}m{1.5cm}}
\toprule
 & 1 & 1S & 2 & 2S & 3 & 4A & overall\\
\midrule
Reports Words & 240.1$\pm$45.7 & 243.5$\pm$2.1 & 291.6$\pm$56.4 & 316.4$\pm$44.3 & 338.0$\pm$0 & 298.3$\pm$56.5 & 280.8$\pm$56.9 \\
Translation Words & 190.9$\pm$43.4 & 276.0$\pm$17.0 & 206.9$\pm$49.0 & 224.4$\pm$111.6 & 155.0$\pm$0 & 210.5$\pm$33.6 & 205.8$\pm$53.9 \\
\bottomrule
\end{tabular}
\end{table}

\begin{table}[!t]
\caption{Comparison of the Brain MRI screening reports and their ChatGPT
translations.\label{table4}}%
\centering
\begin{tabular}{c | c c c | c}
\toprule
 & no mats & stable & worsening & overall \\
\midrule
Reports Words & 158.8$\pm$17.5 & 228.1$\pm$49.1 & 344.9$\pm$83.5 & 256.5$\pm$89.2 \\
Translation Words & 161.7$\pm$31.4 & 198.3$\pm$44.7 & 227.2$\pm$44.4 & 202.5$\pm$47.5 \\
\bottomrule
\end{tabular}
\end{table}

\subsection{Evaluation of ChatGPT translations by radiologists}
We invited two radiologists to evaluate the quality of the translated reports. Evaluation was based on the three metrics: the number of places with information
lost, the number of places with information misinterpreted, and the overall score. The overall score was given based on the 5-point system in which a score of 5 
indicates the best quality while a score of 1 means the worst quality.

Table~\ref{table5} lists statistics of radiologists' evaluation results. It can be found that ChatGPT performed well on both chest CT and brain MRI scan reports. There is only 
0.097 places of information missing and 0.032 places of incorrect information on average per chest CT report, which means once in every 10.3 translated reports and once in every 31.3 translated reports respectively. Among all the translated chest CT reports, 76\% results are rated with an overall score of 5. Regarding the brain MRI scan report translations, there are 5\% results showed information missing, and there are an average 0.066 places of information missing per report. Meanwhile, 9\% of translated reports are with incorrect information, and there are an average 0.092 places of incorrectness per report. 37\% and 32\% of all brain MRI scan results are rated with an overall score of 4 and 5 respectively. Overall, the average numbers of information missing and incorrectness for all results are 0.080 and 0.065 respectively, with a frequency of roughly once in every 12.5 and 15.4 reports respectively. The average overall score of all results is 4.268, in which 27\% and 52\% of all results are rated with an overall score of 4 and 5 respectively.

\begin{table}[!t]
\caption{Radiologists evaluation results.\label{table5}}%
\centering
\begin{tabular}{c | c c c}
\toprule
 & information missing & incorrect information & overall score \\
\midrule
Chest CT & 0.097 & 0.032 & 4.645 \\
Brain MRI & 0.066 & 0.092 & 3.961 \\ \hline
Overall & 0.080 & 0.065 & 4.268 \\
\bottomrule
\end{tabular}
\end{table}

\subsection{Evaluation of ChatGPT-generated suggestions}
When giving suggestions for either patients and healthcare providers, ChatGPT claimed that they could not provide medical advice or treatment at the moment. However, it would provide general suggestions for patients or healthcare providers. We conducted statistical analysis on the ChatGPT-provided suggestions. According to Tables~\ref{table6} and~\ref{table7}, the suggestions for patients and healthcare providers are highly relevant. For example, for the suggestions based on chest CT reports, the most frequently given suggestions for patients and healthcare providers include "follow up with doctors" and "communicate the findings clearly to patient", respectively. For about 37\% of all cases, ChatGPT provided specific suggestions based on findings in the radiology report. Taking a brain MRI report as an example, with an finding on paranasal sinus disease. It is stated in the report that "Paranasal sinuses: Air-fluid levels within maxillary sinuses." ChatGPT gave the following suggestions to the patient and healthcare provider respectively: "Manage sinus symptoms: The report notes that there is air-fluid in the patient's maxillary sinuses (paranasal sinus disease). The patient may want to discuss with their healthcare provider about how to manage any symptoms related to this" and "Evaluate sinus symptoms: The report notes the presence of air-fluid in the patient's maxillary sinuses (paranasal sinus disease). As such, it may be appropriate to evaluate the patient for any symptoms related to this and determine if any treatment or management is necessary".

\begin{table}[!t]
\caption{General suggestions based on the chest CT reports\label{table6}}%
\centering
\begin{tabular}{c | c}
\toprule
Suggestion for a patient &  Frequency (\%) \\
\midrule
Follow up with doctors & 100\% \\
Follow-up with recommended appointments & 100\% \\
Quit smoking & 98\% \\
Maintain a healthy lifestyle & 92\% \\ \hline \hline
Suggestion for a healthcare provider  &  Frequency (\%) \\ \hline
Communicate the findings clearly to the patient & 100\% \\
Schedule follow-up appointments & 100\% \\
Encourage smoking cessation & 98\% \\
Encourage a healthy lifestyle & 65\% \\
Consider referral to a specialist & 40\% \\
Monitor the nodule as recommended & 39\% \\
Document the results in the patient's medical record & 18\% \\
Review report thoroughly & 10\% \\

\bottomrule
\end{tabular}
\end{table}

\begin{table}[!t]
\caption{General suggestions based on brain MRI reports.\label{table7}}%
\centering
\begin{tabular}{c | c}
\toprule
 Suggestion for a patient &  Frequency (\%) \\
\midrule
Follow-up with recommended appointments & 100\% \\
Follow up with doctors & 99\% \\
Maintain a healthy lifestyle & 97\% \\
Monitor symptoms and report any changes to a healthcare provider & 42\% \\ \hline \hline
Suggestion for a healthcare provider & Frequency (\%) \\ \hline
Communicate the findings clearly to the patient & 100\% \\
Schedule follow-up appointments & 97\% \\
Consider referral to a specialist & 80\% \\
Comprehensive treatment plan & 53\% \\
Evaluate the patient's overall health & 36\% \\
Review report thoroughly & 32\% \\
Additional imaging & 28\% \\
Encourage a healthy lifestyle & 17\% \\
\bottomrule
\end{tabular}
\end{table}

\subsection{Robustness of ChatGPT's translations}
It is found that ChatGPT's translation is not unique for any given radiology report, with different lengths of reorganized paragraphs and flexible choices of alternative words. Hence, it is necessary to investigate the randomness of ChatGPT's responses. We collected 10 translations from the same chest CT radiology report and investigated each translated report. We first split the original radiology report into 25 key information points, and then evaluated the correctness and completeness of each corresponding point in every translated report in a point-by-point fashion. Our
results on the chest CT radiology reports are shown in Table~\ref{table8}, where "Good" means that information was clearly translated, "Missing" indicates that the information point was completely lost in the translation, "Inaccurate" stands for only partial information kept in the translated report, and "Incorrect" shows ChatGPT's misinterpretation of the original radiology report. The overall good translation accounts for 55.2\% of all translated points, and there are 19.2\%, 24.8\%, and 0.8\% information points being completely omitted, partially translated, and misinterpreted, respectively. Notably, for the translation of lung nodule findings, all 10 translations only mentioned the stable status of existing nodules compared with the previous screening, and failed to provide detailed information such as the precise position of each nodule and the size of each nodule. As a result, we consider that all lung nodule findings  were inaccurately translated. When "no new nodules" were mentioned in the original report, only one translation reflected that point, and the other nine translations just mentioned the stable status of existing nodules and omit the statement that there is no new nodules in this screening. The only two incorrect information both happened in the translation of patient smoking history. ChatGPT falsely translated 30 pk-yr (30 packs a year) into 30 years. ChatGPT sometimes neglected minor problems mentioned in the original report. The lung finding of "mild emphysema with minor central bronchial wall thickening bilaterally" was only translated into mild emphysema in most of the translations, and the other minor finding of "normal caliber thoracic aorta with minor
atherosclerotic change" was neglected in nine of the ten translations.

\begin{table}[!t]
\caption{Statistics of 10 repeated translations of a chest CT report.\label{table8}}%
\centering
\begin{tabular}{m{0.3cm}|>{\centering\arraybackslash}m{2.4cm}|m{6cm}|>{\centering\arraybackslash}m{1.2cm} >{\centering\arraybackslash}m{1.2cm} >{\centering\arraybackslash}m{1.2cm} >{\centering\arraybackslash}m{1.2cm}}
\toprule
\multicolumn{3}{c}{} & Good & Missing & Inaccurate & Incorrect \\
\midrule
\multicolumn{2}{c|}{\multirow{2}{*}{Description}} & Lung CT screen without cWontrast & 6 & 4 & - & - \\ \cdashline{3-7}
\multicolumn{2}{c|}{} & Scanned on Feb 13, 2023 & 2 & 8 & - & - \\
\hline
\multicolumn{2}{c|}{\multirow{2}{*}[-0.7em]{Indication}} & Lung cancer screening & 10 & - & - & - \\ \cdashline{3-7}
\multicolumn{2}{c|}{} & Patient who has smoked 30 or more packs per year & 4 & - & 4 & 2 \\
\hline
\multicolumn{2}{c|}{Comparison} & Feb 11, 2022 & 2 & 8 & - & - \\
\hline
\multicolumn{2}{c|}{Technic} & Low dose axial CT, “as low as reasonably achievable” protocol & 10 & - & - & - \\
\hline
\multirow{15}{*}[-7em]{\begin{sideways}Findings\end{sideways}} & \multirow{6}{*}[-2em]{Lung Nodules} & Lung nodule 1: nodule in right upper lobe, 4.9x3.4mm, stable & - & - & 10 & - \\ \cdashline{3-7}
 & & Lung nodule 2: pleura-based nodule in right middle lobe, 4.6mm, stable & - & - & 10 & - \\ \cdashline{3-7}
 & & Lung nodule 3: nonsolid round nodule in right lower lobe, 4.2mm, stable & - & - & 10 & - \\ \cdashline{3-7}
 & & Lung nodule 4: nonsolid subpleural round nodule in right lower lobe 4.6mm, stable & - & - & 10 & - \\ \cdashline{3-7}
 & & Lung nodule 5: subpleural nodule in right lower lobe, right lower lobe, 3mm, stable & - & - & 10 & - \\ \cdashline{3-7}
 & & No new nodules & 1 & 9 & - & - \\ \cline{2-7}
 & \multirow{2}{*}[-1.4em]{Lung} & Linear atelectasis and/or scarring in the right upper lobe, right middle lobe, lingula, and left lower lobe is mild  & 10 & - & - & - \\ \cdashline{3-7}
 & & Mild emphysema in the upper lung fields with minor central bronchial wall thickening bilaterally & 2 & - & 8 & - \\ \cline{2-7}
 & \multirow{2}{*}{Pleura} & No pleural thickening or effusion & 10 & - & - & - \\ \cdashline{3-7}
 & & No pneumothorax & 10 & - & - & - \\ \cline{2-7}
 & \multirow{2}{*}{Heart} & Heart size normal & 10 & - & - & - \\ \cdashline{3-7}
 & & No pericardial effusion & 10 & - & - & - \\ \cline{2-7}
 & Coronary Artery Calcification & None & 10 & - & - & - \\ \cline{2-7}
 & Mediastinum/ Hilum/Axillla & No adenopathy & 8 & 2 & - & - \\ \cline{2-7}
 & Other & Normal caliber thoracic aorta with minor atherosclerotic change & 1 & 9 & - & - \\
 \hline
\multirow{3}{*}[-1.4em]{\begin{sideways}Conclusion\end{sideways}} & Overall Lung-RADS Category & 2-benign appearance or behavior & 10 & - & - & - \\  \cline{2-7}
 & Based on Lesion & ID multiple right-sided pulmonary nodules largest in the right upper lobe measuring 4.9 mm & 2 & 8 & - & - \\ \cline{2-7}
 & Management recommendation & Continue annual screening with low dose CT in 12 months, Feb 2024 & 10 & - & - & - \\ \cdashline{2-7}
 \hline
\multicolumn{2}{c|}{S Findings} & Minor sequela of COPD & 10 & - & - & - \\
\bottomrule
\end{tabular}
\end{table}

\subsection{Optimized prompt for improved translation}

It is found that ChatGPT tends to generate different responses given the same input, which reflects uncertainty of the language model. Such a randomness could compromise the quality of translated results. A reason for ChatGPT's a variety of responses is actually the  ambiguity of our prompts. Instead of giving a prompt that only let ChatGPT to translate a radiology report into a plain language version, we optimized our initial prompts to be comprehensive and specific. Our optimized prompt is as follows:

\textit{Please help translate a radiology report into plain language in the following format: 
\begin{itemize}
  \item First paragraph introduces screening description including reason for screening, screening time, protocol, patient background, and comparison date;
  \item Second paragraph talks about specific findings: how many nodules detected, each lung nodule’s precise position and size, findings on lungs, heart, pleura, coronary artery calcification, mediastinum/hilum/axilla, and other findings. Please don’t leave out any information about findings;
  \item Third paragraph talks about conclusions, including overall lung-rads category, management recommendation and follow-up date, based on lesion;
  \item If there are incidental findings, please introduce in the fourth paragraph.
\end{itemize}
}

We collected another 10 ChatGPT plain-language translations of the radiology report, and conducted the same statistical analysis as what was done in the previous subsection. The results are summarized in Table~\ref{table9}. With the much clearer 
prompt, the overall quality of translation was increased from 55.2\% to 77.2\%, and measures on information completely omitted, partially translated, and misinterpreted were reduced to 9.2\%, 13.6\% and 0\% respectively. A good example of using a detailed prompt is the translation of lung nodule 1. In the experiment with a vague prompt, there was no translation keeping the information in the translated report. In contrast, with a detailed prompt, there were 8 out of 10 translations presenting the information on this nodule. For more details, please see our supplementary files.

\begin{table}[!t]
\caption{Statistics on 10 repeated translations of a chest CT report with the optimized prompt.\label{table9}}%
\centering
\begin{tabular}{m{0.3cm}|>{\centering\arraybackslash}m{2.4cm}|m{6cm}|>{\centering\arraybackslash}m{1.2cm} >{\centering\arraybackslash}m{1.2cm} >{\centering\arraybackslash}m{1.2cm} >{\centering\arraybackslash}m{1.2cm}}
\toprule
\multicolumn{3}{c}{} & Good & Missing & Inaccurate & Incorrect \\
\midrule
\multicolumn{2}{c|}{\multirow{2}{*}{Description}} & Lung CT screen without contrast & 3 & 7 & - & - \\ \cdashline{3-7}
\multicolumn{2}{c|}{} & Scanned on Feb 13, 2023 & 10 & - & - & - \\
\hline
\multicolumn{2}{c|}{\multirow{2}{*}[-0.7em]{Indication}} & Lung cancer screening & 10 & - & - & - \\ \cdashline{3-7}
\multicolumn{2}{c|}{} & Patient who has smoked 30 or more packs per year & 10 & - & - & - \\
\hline
\multicolumn{2}{c|}{Comparison} & Feb 11, 2022 & 10 & - & - & - \\
\hline
\multicolumn{2}{c|}{Technic} & Low dose axial CT, “as low as reasonably achievable” protocol & 10 & - & - & - \\
\hline
\multirow{15}{*}[-7em]{\begin{sideways}Findings\end{sideways}} & \multirow{6}{*}[-2em]{Lung Nodules} & Lung nodule 1: nodule in right upper lobe, 4.9x3.4mm, stable & 8 & - & 2 & - \\ \cdashline{3-7}
 & & Lung nodule 2: pleura-based nodule in right middle lobe, 4.6mm, stable & 4 & - & 6 & - \\ \cdashline{3-7}
 & & Lung nodule 3: nonsolid round nodule in right lower lobe, 4.2mm, stable & 3 & - & 7 & - \\ \cdashline{3-7}
 & & Lung nodule 4: nonsolid subpleural round nodule in right lower lobe 4.6mm, stable & 3 & - & 7 & - \\ \cdashline{3-7}
 & & Lung nodule 5: subpleural nodule in right lower lobe, right lower lobe, 3mm, stable & 3 & - & 7 & - \\ \cdashline{3-7}
 & & No new nodules & 4 & 6 & - & - \\ \cline{2-7}
 & \multirow{2}{*}[-1.4em]{Lung} & Linear atelectasis and/or scarring in the right upper lobe, right middle lobe, lingula, and left lower lobe is mild  & 10 & - & - & - \\ \cdashline{3-7}
 & & Mild emphysema in the upper lung fields with minor central bronchial wall thickening bilaterally & 7 & - & 3 & - \\ \cline{2-7}
 & \multirow{2}{*}{Pleura} & No pleural thickening or effusion & 10 & - & - & - \\ \cdashline{3-7}
 & & No pneumothorax & 8 & 2 & - & - \\ \cline{2-7}
 & \multirow{2}{*}{Heart} & Heart size normal & 9 & 1 & - & - \\ \cdashline{3-7}
 & & No pericardial effusion & 10 & - & - & - \\ \cline{2-7}
 & Coronary Artery Calcification & None & 9 & 1 & - & - \\ \cline{2-7}
 & Mediastinum/ Hilum/Axillla & No adenopathy & 9 & 1 & - & - \\ \cline{2-7}
 & Other & Normal caliber thoracic aorta with minor atherosclerotic change & 8 & 2 & - & - \\
 \hline
\multirow{3}{*}[-1.4em]{\begin{sideways}Conclusion\end{sideways}} & Overall Lung-RADS Category & 2-benign appearance or behavior & 8 & - & 2 & - \\  \cline{2-7}
 & Based on Lesion & ID multiple right-sided pulmonary nodules largest in the right upper lobe measuring 4.9 mm & 7 & 3 & - & - \\ \cline{2-7}
 & Management recommendation & Continue annual screening with low dose CT in 12 months, Feb 2024 & 10 & - & - & - \\ \cdashline{2-7}
 \hline
\multicolumn{2}{c|}{S Findings} & Minor sequela of COPD & 10 & - & - & - \\
\bottomrule
\end{tabular}
\end{table}

\subsection{Different prompts on ChatGPT's performance}
We further investigated the effect of prompt engineering on ChatGPT's performance.
Specifically, we changed the first prompt into the following formats:
\begin{itemize}
  \item Please translate a radiology report into plain language for a patient only with high school education.
  \item Please translate a radiology report into plain language for a patient only with undergraduate education.
  \item Please translate a radiology report into plain language for a patient only with graduate education.  
  \item Can you translate a radiology report into plain language that someone without medical training can easily understand?
  \item Your task is to translate a radiology report into plain language that is easy for the average person to understand. Your response should provide a clear and concise summary of the key findings in the report, using simple language that avoids medical jargon. Please note that your translation should accurately convey the information contained in the original report while making it accessible and understandable to a layperson. You may use analogies or examples to help explain complex concepts, but you should avoid oversimplifying or leaving out important details.
\end{itemize}

The first three prompts asked ChatGPT to translate radiology reports according to different education levels respectively. The fourth prompt was designed by ChatGPT based on the prompt "Please design the best prompt for you based on this prompt: Please translate a radiology report into plain language that is easy to understand." The last prompt was designed by a website called promptperfect \cite{promptperfect}. These five prompts are labeled as prompt 1 to prompt 5 respectively.

\begin{figure}
\centering
\includegraphics[width=4in]{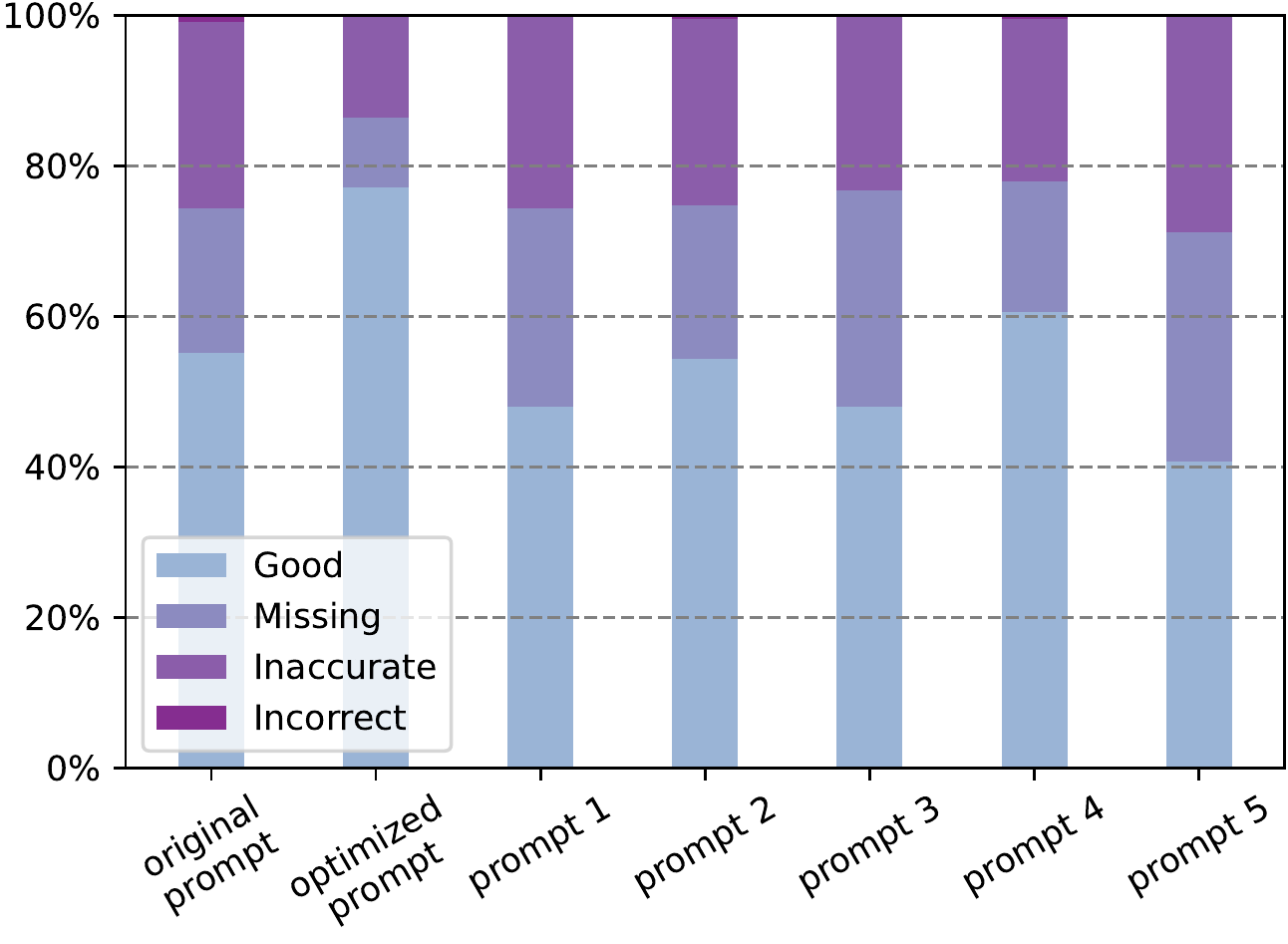}
\caption{Effects of different prompts on ChatGPT's translation performance.} \label{fig1}
\end{figure}

We evaluated ChatGPT's responses with these prompts using the same method as that in the preceding subsection, and compared the new responses with the previous results with the original prompt and the optimized prompt respectively. The results are listed in Fig.~\ref{fig1}. It can be found that all the five further-modified prompts produced results similar to that with the original prompt and far worse results than that with the optimized prompt. In terms of the five modified prompts, the fourth prompt designed by ChatGPT itself  performed slightly better than the other four prompts with a higher good rate and lower missing and inaccurate rates. However, the fourth  prompt still  performed significantly worse than the optimized prompt described in the preceding subsection.

\subsection{ChatGPT's ensemble learning results}

In the above subsections, we asked ChatGPT to generate multiple translated reports with the same prompt and the same radiology report. Here we further investigated ChatGPT's performance via ensemble learning. In each case, we randomly selected 5 translated reports and input into ChatGPT for information integration. We asked ChatGPT to combine all the results to create a single report. Statistics of 10 ensemble learning results are presented in Table~\ref{table10}. Generally, ChatGPT cannot generate significantly better results through ensemble learning. Although it can be found that ChatGPT performed better when combining the results obtained with the original prompt, with a higher good rate and lower missing, inaccurate, and incorrect rates, such an improvement is not significant compared with the improvement by replacing the original prompt with the optimized prompt. In this experiment, the increment of good rate was attributed to reporting more scan details such as pointing out scans were without contrast and more details on patient smoking history. For the integration of results obtained with the optimized prompt, the good rate was declined while missing and accurate rates were increased. This inferior performance mainly resulted from the over-simplification of lung nodule findings and the overlook of minor findings like normal caliber thoracic aorta with minor atherosclerotic change.

\begin{table}[!t]
\caption{Percentage change of ensemble learning versus non-ensemble results.\label{table10}}%
\centering
\begin{tabular}{>{\centering\arraybackslash}m{3cm}|>{\centering\arraybackslash}m{1.5cm}>{\centering\arraybackslash}m{1.5cm}>{\centering\arraybackslash}m{1.5cm}>{\centering\arraybackslash}m{1.5cm}}
\toprule
 & Good & Missing & Inaccurate & Incorrect \\
\midrule
Original prompt & 6.4\% & -0.8\% & -4.8\% & -0.8\% \\
Optimized prompt & -4.4\% & 1.2\% & 3.2\% & 0\% \\
\bottomrule
\end{tabular}
\end{table}

\subsection{Comparison with GPT-4}

On March 14, OpenAI launched its new large language model GPT-4, with an impressive performance on multi-modal tasks~\cite{gpt4tech}. Immediately, we investigated the performance of GPT-4 in this radiology report translation task and compared the results with that from ChatGPT. The experiments were conducted using the original prompt and the optimized prompt using the same methodology as that used in the ChatGPT experiment. According to Table~\ref{table11}, GPT-4 significantly improved the quality of translated reports with higher good rates and lower other rates using both original and optimized prompts. Impressively, GPT-4's results on the original prompt was competitive with ChatGPT using the optimized prompt, and GPT-4 with the optimized prompt almost achieve 100\% good rate, which is exciting!

Similar to ChatGPT, GPT-4 still has some randomness. In the experiment using the optimized prompt, there was a translation failed to follow the provided format. According to the required format, incidental findings should be listed in the fourth paragraph, but GPT-4 showed the incidental finding of chronic obstructive pulmonary disease in the third paragraph along with conclusions.

\begin{table}[!t]
\caption{Comparison of GPT-4 and ChatGPT on the radiology report plain language translation task.\label{table11}}%
\centering
\begin{tabular}{>{\centering\arraybackslash}m{2cm}>{\centering\arraybackslash}m{3cm}|>{\centering\arraybackslash}m{1.5cm}>{\centering\arraybackslash}m{1.5cm}>{\centering\arraybackslash}m{1.5cm}>{\centering\arraybackslash}m{1.5cm}}
\toprule
 & & Good & Missing & Inaccurate & Incorrect \\
\midrule
\multirow{2}{*}{ChatGPT} &  Original prompt & 55.2\% & 19.2\% & 24.8\% & 0.8\% \\
 & Optimized prompt & 77.2\% & 9.2\% & 13.6\% & 0\% \\ \hline
\multirow{2}{*}{GPT-4} &  Original prompt & 73.6\% & 8.0\% & 18.4\% & 0\% \\
 & Optimized prompt & 96.8\% & 1.6\% & 1.6\% & 0\% \\
 \bottomrule
\end{tabular}
\end{table}

\section{Discussions}

Although it is just a first public-convincing step toward artificial general intelligence (AGI), ChatGPT has already demonstrated an amazing capability of organizing words and sentences. It can be used for multiple purposes such as writing news, telling stories, and language translation. In this study, we have evaluated its potential in translating radiology reports into plain language and making suggestions based on the reports. According to our results, ChatGPT has at least three merits for the radiology report translation: conciseness, clarity, and comprehensiveness. For conciseness, ChatGPT deletes redundant words in the original report and summarizes multiple findings in a single sentence. For clarity, commonly-used words will be adopted by ChatGPT to replace complicated medical terminologies so that patients with different education backgrounds can digest the information easily. In terms of comprehensiveness, ChatGPT has a strong ability to understand the original radiology report and to integrate information from different sections of the original report into easily understandable sentences.

Our experiment has also revealed the uncertainty of ChatGPT's responses. Given the same prompt for the same radiology report, ChatGPT will generate distinctive responses  each time, which could result in a variety of translated reports. Such random results are partially inherent to the language model and partially due to the ambiguity of our vague prompt. Our original prompt only gave ChatGPT a generic instruction to translate a radiology report into plain language, and there were no specific instructions on which information was important and should be kept. As a result, ChatGPT tended to generate over-simplified translations and left out important information. Our results suggested a weakness of the current version of ChatGPT: it does not know which information is important and should be kept in a radiology report. Our experiment with an optimized prompt with detailed instructions on which information should be kept has demonstrated that ChatGPT can generate improved results with clearer and more specific instructions. Meanwhile, ChatGPT presents semantic robustness when there is no significant difference between different prompts. According to Fig.~\ref{fig1} and Table~\ref{table10}, when there are no clear instructions on how to preserve information, ChatGPT's performances are similar with multiple semantically similar prompts.

Another interesting finding is that it seems that ChatGPT does not have a built-in template for its generated report translation. Radiology reports usually follow a fixed template so that reports made by different radiologists are presented in a consistent way. Such a consistent template greatly improves the efficiency of radiology report generation and saves time for healthcare providers to digest radiology reports. According to our results, ChatGPT tends to generate results in various formats when a prompt has no format instruction. In some cases, ChatGPT will produce a single-paragraph translation with all findings and conclusions combined together. Compared with translated reports that have multiple paragraphs and present information about screening description, findings, and conclusions in different paragraphs, those single-paragraph translated reports are more difficult for patients to read. Designing a prompt with clear instructions on the format of the translated reports can help ChatGPT generate translations in a consistent format with better readability. For example, the number of paragraphs and the number of words can be added into the prompt to specify the format and the length of the translation.

According to the evaluation results by our radiologists, ChatGPT's translated results are with few missing information and mis-interpretation, alleviating concerns on the reliability of ChatGPT's translation results. Currently, the development of large language models has been very rapid, and new models is frequently released. Taking GPT-4 as an example, it was launched on March 14, 2023 with the ability to deal with multi-modality data like text and images, which performs better on multiple tasks like uniform bar exam than its predecessor GPT-3.5~\cite{gpt4}. Our results on GPT-4 also demontrate the significant improvemnt compared with ChatGPT. The future will be really bright of utilizing large language models for clinical applications.

Despite the potential of ChatGPT in radiology report translation, there remain concerns to be addressed before its deployment in clinical practice. The first concern is that ChatGPT's report translation is still lack of completeness, which may leave out some key points. According to our results, using an optimized prompt can improve completeness, however, the current results are still not perfect. Another concern is the inconsistency or uncertainty of ChatGPT's responses. ChatGPT may give inconsistent translations and present information in variable formats with potential over-simplifications or information losses for the same radiology report with the same prompt. 

In terms of using AI in the healthcare domain, large language models like ChatGPT has  demonstrated its potential. This study is a good example in which radiology reports can be translated into plain language efficiently and effectively, even with
useful suggestions automatically without direct involvement of human experts. In the future, ChatGPT-type systems will be surely and extensively used in healthcare to  provide great assistance such as generating full radiology reports directly from medical images, analyzing treatment options and plans, guiding patients' daily life taking all their medical data into consideration, and providing psychological counseling as needed. 

It is clear that ChatGPT and products like it will greatly impact the way in which medical information is formulated, queried, and shared across patients and healthcare providers. The evidence needed to demonstrate to regulators that such algorithms are safe and effective will depend on their intended use as well as the risks and benefits of such uses. Tools that support communication between healthcare providers and patients are likely to be easier to demonstrate safety over those that have a more direct impact on patient diagnosis and treatment planning. We look forward to seeing the further development of these products and additional evaluations of their performance characteristics for the purpose of regulatory review and confident adoption by users.

\section{Conclusion}
In conclusion, we have investigated the feasibility and utility of applying ChatGPT
in  low-hanging clinical applications specially translation of radiology reports into plain language and recommendations to a patient or a healthcare provider, and conducted experiments to evaluate ChatGPT's performance in this particular clinical task. According to our professional evaluative results, ChatGPT's translations have an overall score of 4.268 in the five-point system (5 for best and 1 for worst) with an average of 0.097 places of information missing and 0.065 places of incorrect information per translation. As far as the uncertainty of ChatGPT's responses is concerned, it is found that ChatGPT's plain language translation tends to over-simplify or over-look some key points, with only 55.2\% key points completely translated when using a vague prompt. Such an uncertainty can be reduced to reach 77.2\% information fully translated when replacing the vague prompt with an optimized prompt. We have further compared ChatGPT with GPT-4 and found that GPT-4 can significantly improve the quality of translated reports. Our study have showed that advanced large language models like ChatGPT and GPT-4 are promising new tools in clinical applications, and an initial translational project should be translation of radiology reports into plain language.


\bibliographystyle{unsrt}  
\bibliography{references}

\end{document}